\documentclass[onefignum,onetabnum]{siamart171218}

\usepackage{lipsum}
\usepackage{amsfonts}
\usepackage{graphicx}
\usepackage{epstopdf}
\usepackage{algorithmic}
\ifpdf
  \DeclareGraphicsExtensions{.eps,.pdf,.png,.jpg}
\else
  \DeclareGraphicsExtensions{.eps}
\fi

\newsiamremark{remark}{Remark}
\newsiamremark{hypothesis}{Hypothesis}
\crefname{hypothesis}{Hypothesis}{Hypotheses}
\newsiamthm{claim}{Claim}

\headers{Batch-efficient Divide-and-Conquer}{Y. Song}

\title{A Short Note on Batch-efficient Divide-and-Conquer Algorithm for EigenDecomposition}

\author{Yue Song\thanks{College of AI, Tsinghua University, China 
  (\email{yue-song@mail.tsinghua.edu.cn}, \url{https://kingjamessong.github.io/}).}
}

\usepackage{amsopn}

\makeatletter
\newcommand*{\addFileDependency}[1]{% argument=file name and extension
  \typeout{(#1)}% latexmk will find this if $recorder=0 (however, in that case, it will ignore #1 if it is a .aux or .pdf file etc and it exists! if it doesn't exist, it will appear in the list of dependents regardless)
  \@addtofilelist{#1}% if you want it to appear in \listfiles, not really necessary and latexmk doesn't use this
  \IfFileExists{#1}{}{\typeout{No file #1.}}% latexmk will find this message if #1 doesn't exist (yet)
}
\makeatother

\ifpdf
\hypersetup{
  pdftitle={A Short Note on Batch-efficient Divide-and-Conquer Algorithm for EigenDecomposition},
  pdfauthor={Y. Song}
}
\fi

\def\mD{{\mathbf{D}}}

\def\mI{{\mathbf{I}}}

\def\mQ{{\mathbf{Q}}}

\def\mT{{\mathbf{T}}}

\begin{document}

\maketitle

% REQUIRED
\begin{abstract}
  EigenDecomposition (ED) is at the heart of many computer vision algorithms and applications. One crucial bottleneck limiting its usage is the expensive computation cost, particularly for a mini-batch of matrices in deep neural networks. Our previous work proposed a dedicated QR-based ED algorithm for batched small matrices (dim${<}32$). This short paper targets the limitation and proposes a batch-efficient Divide-and-Conquer based ED algorithm for larger matrices. The numerical test shows that for a mini-batch of matrices whose dimensions are smaller than $64$, our method can be much faster than the Pytorch SVD function. The Pytorch implementation is available at: \href{https://github.com/KingJamesSong/BatchED}{https://github.com/KingJamesSong/BatchED}.
\end{abstract}

\section{Introduction}

\begin{figure}[htbp]
    \centering
    \includegraphics[width=0.8\linewidth]{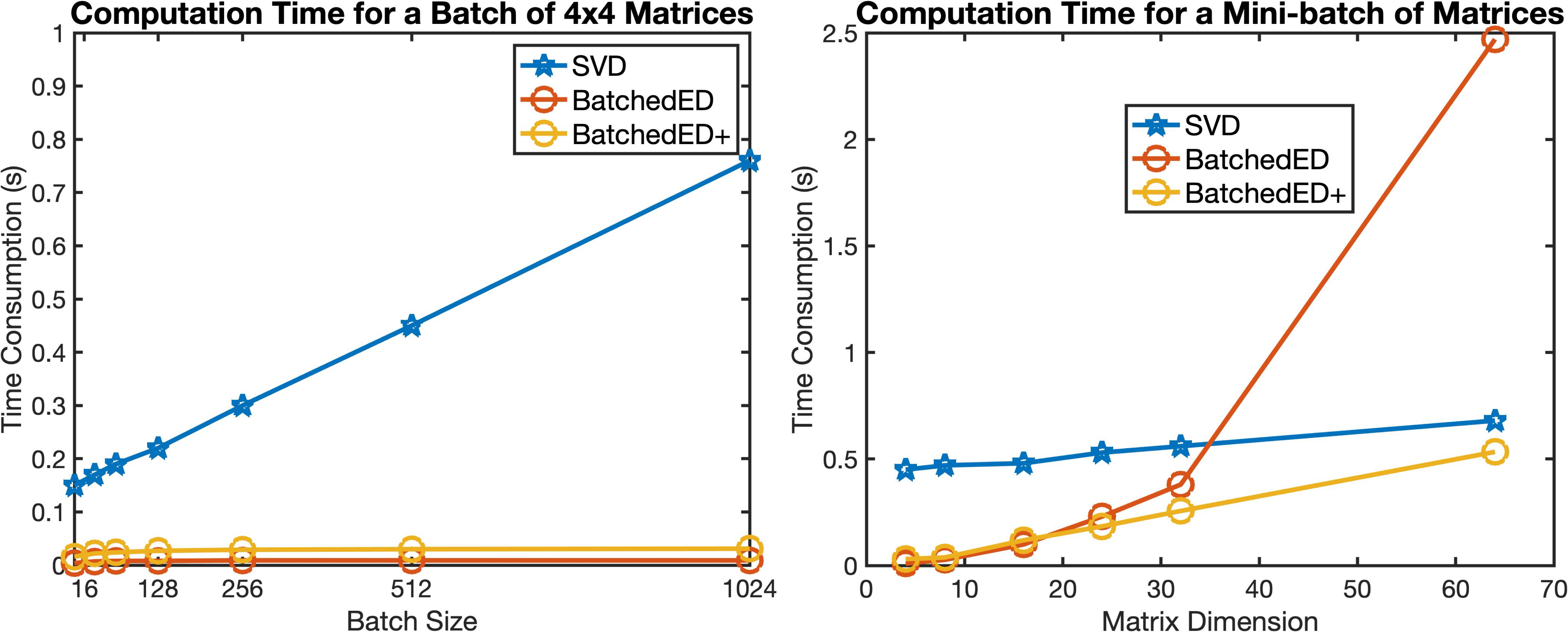}
    \caption{The speed comparison of our BatchedED and BatchedED+ against the \textsc{torch.svd}. (\emph{Left}) Time consumption for a mini-batch of $4{\times}4$ matrices with different batch sizes. (\emph{Right}) Time consumption for matrices with batch size $512$ but in different matrix dimensions. Our proposed BatchedED+ is more efficient for larger matrices. }
    \label{fig:cover}
\end{figure}

The EigenDecomposition (ED) explicitly factorizes a matrix into the eigenvalue and eigenvector matrix, which serves as a fundamental tool in computer vision and deep learning. Recently, many algorithms integrated the SVD as a meta-layer into their models to perform some desired spectral transformations~\cite{song2022fast2}. The problem setup of the ED in computer vision is quite different from other fields. In other communities such as scientific computing, batched matrices rarely arise and the ED is usually used to process a single matrix. However, in deep learning and computer vision, the model takes a mini-batch of matrices as the input, which raises the requirement for an ED solver that works for batched matrices efficiently. Moreover, the differentiable ED works as a building block and needs to process batched matrices millions of times during the training and inference. This poses a great challenge to the efficiency of the ED solver and could even stop people from adding the ED meta-layer in their models due to the huge time consumption (see Fig.~\ref{fig:cover}).

Our previous work~\cite{song2022batch} proposed a QR-based batch-efficient algorithm dedicated to small matrices (dim${<}32$). However, the quadratic time complexity $O(n^{3})$ of QR iterations would limit the application on large matrices. To alleviate this issue, this short paper extends the algorithm and propose a Divide-and-Conquer (DC) based batch-efficient ED method for larger matrices (dim${<}64$). Our method casts the classical DC algorithm into a constrained optimization problem, \emph{i.e.,} solving secular equations with interleaved eigenvalue constraint. The joint use of hybrid-section and Halley's method are adopted to efficiently localize the eigenvalues. Moreover, the progressive batch removal is proposed to gradually reduce the computational burden. The numerical tests demonstrate that, for batched matrices, our Pytorch implementation is consistently much faster than the default SVD routine (see also Fig.~\ref{fig:cover}).

%\section{Main results}

\section{Algorithm}

This section presents our batch-efficient DC algorithm. For the batched Householder reflection, please refer to our previous work~\cite{song2022batch} for detail.

\subsection{Batch-efficient Divide Step}

The divide step aims at dividing the matrix recursively until the matrix size is small enough and the processing time sufficiently is cheap. Consider a tri-diagonal matrix $\mT$:
\begin{equation}
    \mT = \begin{bmatrix}
    a & b & 0 & 0 \\
    b & c & \beta & 0 \\
    0 & \beta & e & f \\
    0 & 0 & f & g \\
    \end{bmatrix} = \begin{bmatrix} 
      \multicolumn{2}{c}
      {\raisebox{\dimexpr\normalbaselineskip-2.0\height}[0pt][0pt]
        {\scalebox{1.3}{$\mT_{1}$}}} & 0 & 0 \\
        & & \beta & 0 \\
        0 & \beta &  \multicolumn{2}{c}
      {\raisebox{\dimexpr\normalbaselineskip-2.0\height}[0pt][0pt]
        {\scalebox{1.3}{$\mT_{2}$}}}\\
        0 & 0 & &\\
    \end{bmatrix}
\end{equation}
The matrix can be partitioned as:
\begin{equation}
\begin{aligned}
    \mT &= \begin{bmatrix} 
      \multicolumn{2}{c}
      {\raisebox{\dimexpr\normalbaselineskip-1.5\height}[0pt][0pt]
        {\scalebox{1.3}{$\hat{\mT}_{1}$}}} & 0 & 0 \\
        & & 0  & 0 \\
        0 & 0 &  \multicolumn{2}{c}
      {\raisebox{\dimexpr\normalbaselineskip-1.5\height}[0pt][0pt]
        {\scalebox{1.3}{$\hat{\mT}_{2}$}}}\\
        0 & 0 & &\\
    \end{bmatrix} + \begin{bmatrix}
    0 & 0 & 0 & 0\\
    0 & |\beta| & \beta & 0\\
    0 & \beta & |\beta| & 0\\
    0 & 0 & 0 & 0\\
    \end{bmatrix} \\
    &= \begin{bmatrix}\hat{\mT}_{1} & \\
     & \hat{\mT}_{2}
    \end{bmatrix} + |\beta| \mathbf{v}\mathbf{v}^{T}
    \label{eq:two_eig_rankone}
\end{aligned}
\end{equation}
where $\mathbf{v}$ is a vector defined by $\mathbf{v}{=}[0,\pm1,1,0]$, and $\hat{\mT}_{1}$ and $\hat{\mT}_{2}$ are two smaller tri-diagonal matrices defined by:
\begin{equation}
  \hat{\mT}_{1}=\begin{bmatrix}
  a & b \\
  b & c-|\beta|\end{bmatrix}, \hat{\mT}_{2}=\begin{bmatrix}
  e-|\beta| & f \\
  f & g \end{bmatrix}
\end{equation}
Now the problem becomes solving two smaller eigenvalue problem plus a rank-one modification. This kind of partition continues until the matrix is irreducible, \emph{i.e.,} the matrix is of size $2{\times}2$. Alternatively, we can first perform the divide step a few times and switch to our batch-efficient QR iterations or the off-the-shelf function \textsc{torch.svd}. For the detailed combination, we will discuss it in Sec.~\ref{sec:comb}.

Notice that the divide step naturally raises the need for the batch-efficient algorithm. Consider the original tri-diagonal matrix $\mT$ of size $C{\times}C$. The recursive partition generates a mini-batch of matrices $R{\times}\frac{C}{R}{\times}\frac{C}{R}$, where $R$ denotes the partition times. For batched matrices of size $B{\times}C{\times}C$, the partitioned matrices will be of size $BR{\times}\frac{C}{R}{\times}\frac{C}{R}$. Therefore, the batch-efficient algorithm can be very advantageous in processing batched matrices.

\subsection{Combination with Other Eigensolvers}
\label{sec:comb}

After partitioning the original matrices into many smaller matrices, there are multiple options to process these small matrices. Now we introduce each of them in detail.

\noindent \textbf{Off-the-shelf \textsc{torch.svd} (LAPACK's SVD routine).} One method to process batched matrices is via Pytorch self-contained LAPACK's SVD routine. The SVD routine can be directly called to process the partitioned matrices.

\noindent \textbf{Batch-efficient QR iteration.} We can also switch to our batched QR~\cite{song2022batch}. 

\noindent \textbf{Givens Rotation for $2{\times}2$ Block.} The most convenient step is to partition the matrix until it is not reducible anymore, \emph{i.e.,} matrices of size $2{\times}2$.

\subsection{Batch-efficient Conquer Step}

The process of emerging the small matrices into the matrix of original size is called conquer step. Now we illustrate the detailed process and our modification to make the algorithm batch-efficient. 

\begin{figure}
    \centering
    \includegraphics[width=0.6\linewidth]{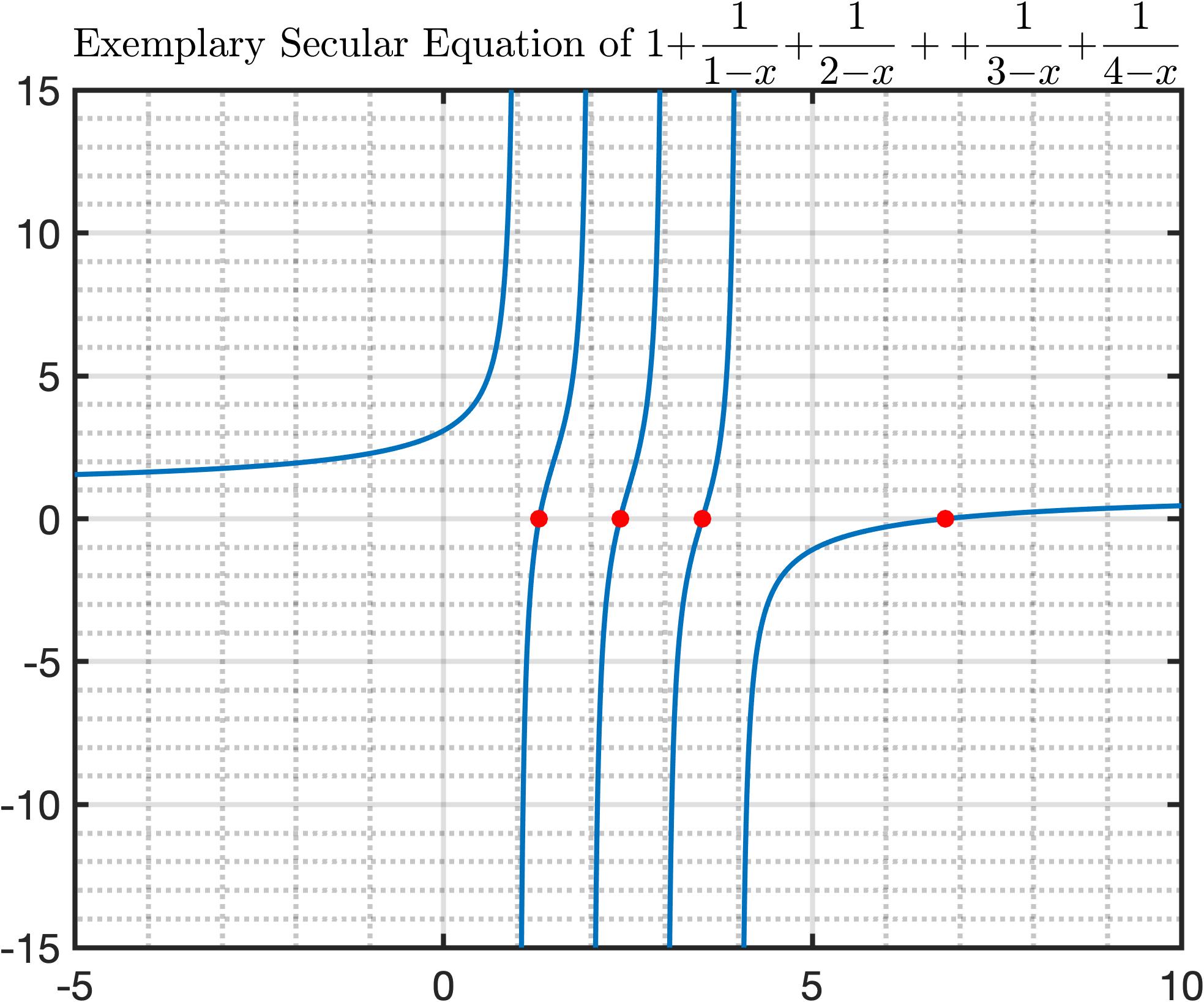}
    \caption{Exemplary secular equations with roots highlighted. }
    \label{fig:secular}
\end{figure}

\subsubsection{Secular Equations}

Suppose we have the spectral decomposition of $\hat{\mT}_{1}{=}\mQ_{1}\mD_{1}\mQ_{1}^{T}$ and $\hat{\mT}_{2}{=}\mQ_{2}\mD_{2}\mQ_{2}^{T}$. The problem in~\cref{eq:two_eig_rankone} can be reformulated as:
\begin{equation}
\begin{aligned}
    \mT &= \begin{bmatrix}\mQ_{1}\mD_{1}\mQ_{1}^{T} & \\
     & \mQ_{2}\mD_{2}\mQ_{2}^{T}\\
    \end{bmatrix} + |\beta| \mathbf{v}\mathbf{v}^{T} \\
    &= \begin{bmatrix}
    \mQ_{1} & \\
    & \mQ_{2} \\
    \end{bmatrix} \Big( \mD+|\beta|\mathbf{z}\mathbf{z}^{T}\Big)
    \begin{bmatrix}
    \mQ_{1}^{T} & \\
    & \mQ_{2}^{T} \\
    \end{bmatrix}
\end{aligned}
\end{equation}
where $\mD$ is the diagonal matrix comprising the eigenvalues of $\mT_{1}$ and $\mT_{2}$, and $z$ is computed as:
\begin{equation}
    \mathbf{z}=\begin{bmatrix}
    \mQ_{1} & \\
    & \mQ_{2} \\
    \end{bmatrix}\mathbf{v}=\begin{bmatrix}
    \pm the\ last\ row\ of\ \mQ_{1} \\
    the\ first\ row\ of\ \mQ_{2} \\
    \end{bmatrix}
\end{equation}
Finding the eigenvalues of $\mD+|\beta|\mathbf{z}\mathbf{z}^{T}$ is equivalent to finding the roots of the characteristic polynomial:
\begin{equation}
    \det{(\mD {+} |\beta|\mathbf{z}\mathbf{z}^{T} {-}\lambda\mI)}{=}\det{((\mD{-}\lambda\mI)(\mI {+} |\beta|(\mD{-}\lambda)^{-1}\mathbf{z}\mathbf{z}^{T}))}
\end{equation}
Since $(\mD-\lambda\mI)$ is non-singular, we have:
\begin{equation}
    \det{(\mI + |\beta|(\mD-\lambda)^{-1}\mathbf{z}\mathbf{z}^{T})}=0
\end{equation}
This relation holds for any eigenvalue. Then deriving the eigenvalue requires solving the following secular equations:
\begin{equation}
    f(\lambda)=1+|\beta|\sum_{i=1}^{n} \frac{z_{i}^2}{d_{i}-\lambda}=0
    \label{secular}
\end{equation}
Fig. depicts the distributions of poles and zeros when $\beta{>}0$. As can be seen, the distribution of the poles ($d_{i}$) and the zeros ($\lambda_{i}$) satisfies the interlacing property:
\begin{equation}
\begin{gathered}
    d_{n}<\lambda_{n}<d_{n-1}<\lambda_{n-1}<\dots<d_{1}<\lambda_{1}%\ \ if\  \beta>0;\\
    %\lambda_{n}<d_{n}<\lambda_{n-1}<<d_{n-1}<\dots<\lambda_{1}<d_{1}\ \ if\  \beta<0.
    \label{interlace}
\end{gathered}
\end{equation}
Notice that the largest eigenvalue $\lambda_{1}$ when $\beta>0$ does not have an upper bound, and the smallest eigenvalue $\lambda_{n}$ does not have a lower bound. Since the matrix is positive semi-definite, we have $\lambda_{n}{>}0$ and the lower bound of $\lambda_{n}$ can be defined. For the upper bound of $\lambda_{1}$, we have:
\begin{equation}
    ||\mT||_{\rm F} =\sqrt{\sum T_{ij}^2}= \sqrt{\sum_{i=1}^{n}\lambda_{i}^2}\geq\lambda_{1}
\end{equation}
where the equality only holds when $\mT$ is a rank-one matrix (\emph{i.e.,} $\lambda_{2}{=}{\dots}{=}\lambda_{n}{=}0$). Then we can add the bound to~\cref{interlace} as:
\begin{equation}
    \begin{gathered}
    d_{n}<\lambda_{n}<d_{n-1}<\lambda_{n-1}<\dots<d_{1}<\lambda_{1}\leq||\mT||_{\rm F}%\ \ if\  \beta>0;\\
    %0\leq\lambda_{n}<d_{n}<\lambda_{n-1}<d_{n-1}<\dots<\lambda_{1}<d_{1}\ \ if\  \beta<0.
    \label{interlace_bounded}
\end{gathered}
\end{equation}

Traditional Divide-and-Conquer algorithms use different solutions to obtain the desired eigenvalues. However, these techniques are not efficient in processing batched matrices. 

To attain a batch-efficient conquer step, we need to cast the problem into a batch-efficient constrained optimization problem. Let $\vec{\lambda}{\in}\mathbb{R}^{B{\times}C}$ denotes the eigenvalues of the batched matrices. Then each eigenvalue in $\Vec{\lambda}$ should satisfy:
\begin{equation}
    Solve\ f(\lambda)=1+\beta\sum_{i=1}^{n} \frac{z_{i}^2}{d_{i}-\lambda}=0 \ \ s.t.~\cref{interlace_bounded}
\end{equation}
The problem can be solved using any zero-order or first-order optimization method (\emph{e.g.,} Bisection method or Newton's method). However, as can be observed from Fig. , the secular equation has flat curve in the middle region but has sharp line near the bounds. Solely using one optimization method could cause the slow convergence and expensive computation. Therefore, it is appropriate to first narrow the eigenvalue range and then switch to the gradient-based optimization method. We propose to use the Hybrid-section method to narrow the eigenvalue range and then switch to second-order optimization method to accurate localize the eigenvalue.  

\subsubsection{Hybrid-section Method to Narrow the Range}

Our hybrid-section method consists of joint usage of bi-section method and multi-section method. Such a usage can fast define the narrow eigenvalue range. 

We first obtain an initial estimate of the eigenvalue by:
\begin{equation}
    h_{i}=\frac{d_{i}+d_{i+1}}{2}
\end{equation}
where $h_{i}$ is the estimate of the corresponding eigenvalue $\lambda_{i}$. Let $U_{i}$ and $L_{i}$ denotes the upper and lower bound of $h_{i}$, respectively. The bisection method updates the upper and lower bounds as:
\begin{equation}
\begin{gathered}
    U_{i}=\begin{cases}U_{i} & f(h_{i})<0\\
    h_{i} &  f(h_{i})>0
    \end{cases}; 
    L_{i}=\begin{cases}L_{i} & f(h_{i})>0\\
    h_{i} &  f(h_{i})<0
    \end{cases}
    \label{eq:update}
\end{gathered}
\end{equation}
There follows the update on the estimate as:
\begin{equation}
    h_{i}=\frac{U_{i}+L_{i}}{2}
    \label{bisection}
\end{equation}
The bi-section method can estimate the eigenvalue range but might not exploit the property of secular equations. Since $\beta\mathbf{z}\mathbf{z}^{T}$ is a rank-one update on the diagonal matrix $\mathbf{D}$, the eigenvalues of $\mathbf{D}$ should be minor updated. Thus, in most cases, the eigenvalues $\mathbf{D}+\beta\mathbf{z}\mathbf{z}^{T}$ should be closer to the lower bound, \emph{i.e.,}the eigenvalues of $\mathbf{D}$, instead of the upper bound. Intuitively, the multi-section method should be more efficient than the bisection method. Let $k$ denotes the number of divided sections. After one iteration of bi-section method, we switch to the multi-section method~\cref{bisection} as:
\begin{equation}
    h_{i}= L_{i} + \frac{U_{i}-L_{i}}{k}
\end{equation}
The updating process follows as done in~\cref{eq:update}. The hybrid-section method is performed several times until the estimation range is sufficiently small. The termination criterion is defined as:
\begin{equation}
    U_{i}-L_{i}< \epsilon
\end{equation}
where $\epsilon$ is a small value such that the following optimization method can quickly converge to the eigenvalue.

\subsubsection{Halley's Method to Accurately Localize Eigenvalues}

Now we switch to the higher-order optimization method to locate the eigenvalues. The optimization method involves the gradient of the secular equations in~\cref{secular}. The first-order gradient is computed as:
\begin{equation}
    f'(h)=\beta \sum_{i=1}^{n} \frac{z_{i}^2}{ (d_{i}-h)^2 }
\end{equation}
After having a reasonable estimate of $h_{i}$, Newton's method gives the iterative update by:
\begin{equation}
    h_{i}^{next}=h_{i}^{prev} - \frac{f(h_{i}^{prev})}{f'(h_{i}^{prev})}
\end{equation}
The Newton's method continues until the relative change $||h_{i}^{next}{-}h_{i}^{prev}||_{\rm F}$ is below a certain tolerance. However, the first-order Newton's method not scale well to large matrices. Since the closed intervals of large matrices are typically very large, the intervals after section method are not in the tangent neighborhood of eigenvalues. The Newton's method, which is locally convergent, thus have poor performances. 

To address this issue, we propose to use the Halley' method, which has cubic convergence speed, to compute the zeros of the secular equations. The Halley's method integrates the second-order derivative into the iterative root-finding update as:
\begin{equation}
    h_{i}^{next}=h_{i}^{prev} - \frac{2f(h_{i}^{prev})f'(h_{i}^{prev})}{2(f'(h_{i}^{prev}))^2-f(h_{i}^{prev})f''(h_{i}^{prev})}
    \label{halley}
\end{equation}
The second-order derivative of the secular equations can be easily computed by:
\begin{equation}
    f''(h)=2\beta \sum_{i=1}^{n} \frac{z_{i}^2}{ (d_{i}-h)^3 }
\end{equation}

\subsubsection{Progressive Batch Dimension Shrinkage}

During the iterations of the Halley's method, we can progressively reduce the batch dimension by checking the convergence of each matrix. That is, when the secular equation of a certain matrix is close to zero, \emph{i.e.,} $|f(\Vec{\lambda}[i])|\leq\epsilon$ where $i$ is the matrix index and $\epsilon$ is the error tolerance, we can perform the shrinkage as:
\begin{equation}
    \vec{\lambda}\in\mathbb{R}^{B{\times}C} \rightarrow \vec{\lambda}\in\mathbb{R}^{(B-1){\times}C}
\end{equation}
In this way, the dimension of the eigenvalue vector is gradually reduced during the optimization process.

\subsubsection{Stable Eigenvector Calculation}

When the eigenvector is required, we have:
\begin{equation}
    \mathbf{q}_{i}=(\lambda_{i}\mI -\mD)^{-1}\mathbf{z}=(\frac{z_{1}}{d_{1}-\lambda_{i}},\dots,\frac{z_{n}}{d_{n}-\lambda_{i}})^{T} 
    \label{vec_unnorm}
\end{equation}
However, this method cannot cannot keep the compact orthogonal form of the eigenvector
\begin{equation}
\begin{aligned}
    \mathbf{q}_{i} &= \frac{(\lambda_{i}\mI -\mD)^{-1}\mathbf{z}}{||(\lambda_{i}\mI -\mD)^{-1}\mathbf{z}||} \\ 
    &= (\frac{z_{1}}{d_{1}-\lambda_{i}},\dots,\frac{z_{n}}{d_{n}-\lambda_{i}})^{T} \Big/ \sqrt{\sum_{j=1}^{n}\frac{z_{j}^2}{(d_{j}-\lambda_{i})^2}}
    \label{vec_norm}
\end{aligned}
\end{equation}
This approach might be still sensitive to errors. Gu \emph{et al.}~\cite{gu1994stable} propose a backward-stable method to compute the eigenvector. After obtaining the eigenvalues, they first re-compute the entry of $\mathbf{z}$ as:
\begin{equation}
    \hat{z}_{i} = \sqrt{\frac{\prod_{j}(\lambda_{j}-d_{i})}{|\beta|\prod_{j\neq i}(d_{j}-d_{i})}}
\end{equation}
The result has been shown to correspond to the prescribed eigenvalue. Then the eigenvector is computed according to~\cref{vec_unnorm}.

\section{Experimental Results}
This section presents the result of numerical test on speed and error for our proposed BatchedED+ solver.

\subsection{Speed Comparison}

\begin{figure}
    \centering
    \includegraphics[width=0.99\linewidth]{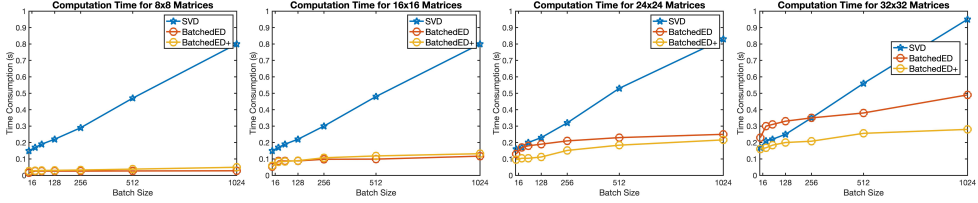}
    \caption{The speed comparison of our BatchedED+ against \texttt{torch.svd} and BatchedED~\cite{song2022batch} for different batch sizes and matrix dimensions. Compared with BatchedED~\cite{song2022batch}, our BatchedED+ is more efficient for larger dimensions.}
    \label{fig:numerical_test}
\end{figure}

%Explain why our BatchedED is inferior to SVD in small batch sizes.
\begin{figure}
    \centering
    \includegraphics[width=0.8\linewidth]{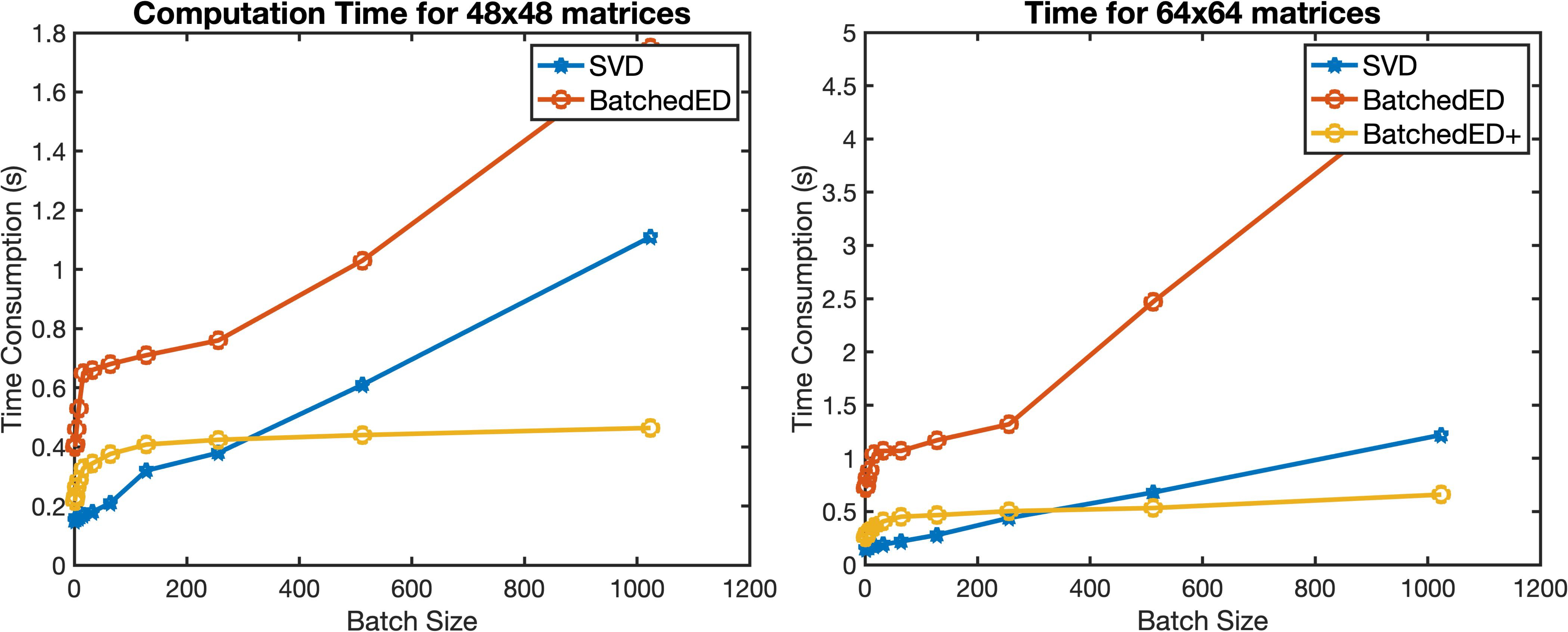}
    \caption{The speed comparison of larger matrices. Our BatchedED+ is more advantagerous when the batch size is also large.}
    \label{fig:num_large}
\end{figure}

Fig.~\ref{fig:numerical_test} compares our BatchedED+ against \texttt{torch.svd} and BatchedED~\cite{song2022batch} in terms of computational speed for different batch sizes and matrix dimensions. When the matrix dimension is small, the speed of our BatchedED+ is slightly inferior than BatchedED~\cite{song2022batch}. However, when it comes to larger matrices, our BatchedED+ becomes more efficient. This point is further verified in Fig.~\ref{fig:num_large}: our BatchedED+ is consistently much more efficient for larger matrices. The underlying reason is due to the fundamental difference between QR and DC algorithms. 

\begin{figure}
    \centering
    \includegraphics[width=0.6\linewidth]{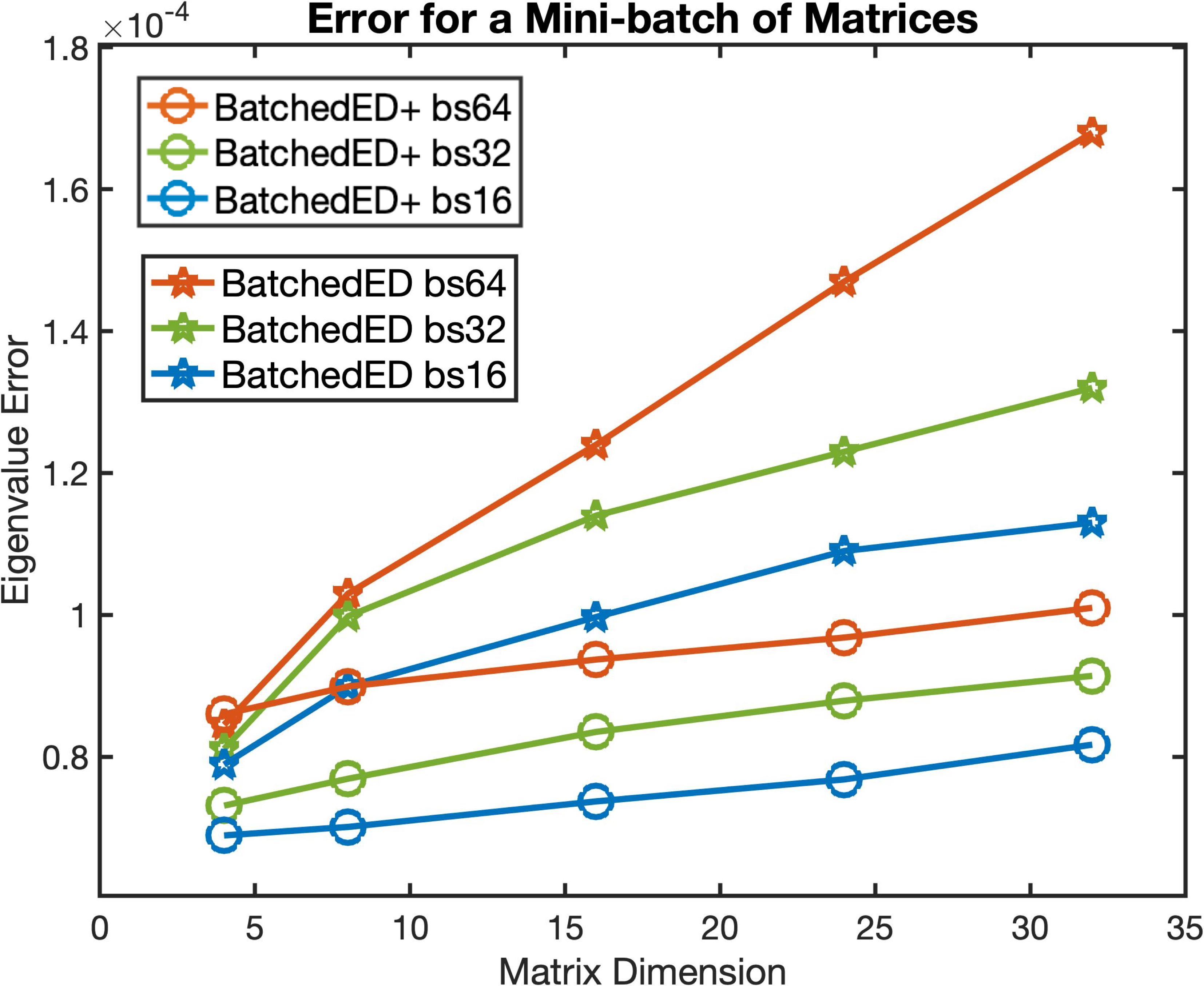}
    \caption{The error of a mini-batch of matrices versus different matrix dimensions.}
    \label{fig:err_dim}
\end{figure}

Fig.~\ref{fig:err_dim} presents the computational error for matrices in different sizes.

\section{Conclusions}

This short note targets the limitation of our previous work~\cite{song2022batch} and extends the idea of batch-efficient ED algorithm to application scenarios of larger matrices. We propose a DC-based ED solver for matrices in larger dimensions. Extensive numerical tests demonstrate that our proposed BatchedED+ is more efficient than BatchedED~\cite{song2022batch} and \texttt{torch.svd} for a mini-batch of large matrices.

\bibliographystyle{siamplain}
\bibliography{references}
\end{document}